%% file: main.tex
\documentclass[10pt,twocolumn,letterpaper]{article}

\usepackage[pagenumbers]{cvpr} % To force page numbers, e.g. for an arXiv version


\definecolor{cvprblue}{rgb}{0.21,0.49,0.74}
\usepackage[pagebackref,breaklinks,colorlinks,allcolors=cvprblue]{hyperref}

\def\paperID{15694} % *** Enter the Paper ID here
\def\confName{CVPR}
\def\confYear{2025}

\title{\modelname: Video VAE with Decoupled Structure and Dynamics
}
\newcommand*\samethanks[1][\value{footnote}]{\footnotemark[#1]}

\author{\textbf{Yuchi Wang$^{1,2}$\thanks{~~Work done during an internship at Microsoft Research Asia.}, ~~Junliang Guo$^{2}$\thanks{~~Junliang Guo and Xu Sun are corresponding authors.}, ~~Xinyi Xie$^{2,3}$,~~Tianyu He$^2$,~~Xu Sun$^1$\samethanks,~~Jiang Bian$^2$ }\\
 $^{1}$Peking University 
   $~^{2}$Microsoft Research Asia
   $~^{3}$CUHK (SZ) \\
   \texttt{wangyuchi@stu.pku.edu.cn, 120040057@link.cuhk.edu.cn, xusun@pku.edu.cn} \\
   \texttt{\{junliangguo, tianyuhe, jiang.bian\}@microsoft.com} \\
   ~\\
\textbf{\url{https://aka.ms/vidtwin}
   }}

\newcommand{\modelname}{VidTwin\xspace}
\newcommand{\lf}{\textit{Structure Latent}\xspace}
\newcommand{\hf}{\textit{Dynamics Latent}\xspace}

\begin{document}

\maketitle
\input{sections/0_abstract}
\input{sections/1_intro}

\input{sections/2_related}

\input{sections/3_method}

\input{sections/4_exp}
\input{sections/5_conclusion}


{
    \small
    \bibliographystyle{ieeenat_fullname}
    \bibliography{main}
}

\input{sections/appendix}

\end{document}

%% file: sections/0_abstract.tex
\begin{abstract}
    Recent advancements in video autoencoders (Video AEs) have significantly improved the quality and efficiency of video generation. In this paper, we propose a novel and compact video autoencoder, \modelname, that decouples video into two distinct latent spaces: Structure latent vectors, which capture overall content and global movement, and Dynamics latent vectors, which represent fine-grained details and rapid movements. Specifically, our approach leverages an Encoder-Decoder backbone, augmented with two submodules for extracting these latent spaces, respectively. The first submodule employs a Q-Former to extract low-frequency motion trends, followed by downsampling blocks to remove redundant content details. The second averages the latent vectors along the spatial dimension to capture rapid motion. Extensive experiments show that \modelname achieves a high compression rate of 0.20\% with high reconstruction quality (PSNR of 28.14 on the MCL-JCV dataset), and performs efficiently and effectively in downstream generative tasks. Moreover, our model demonstrates explainability and scalability, paving the way for future research in video latent representation and generation. 
\end{abstract}

%% file: sections/1_intro.tex
\section{Introduction}
\label{sec:intro}

\begin{figure}
    \centering
    \includegraphics[width=1\linewidth]{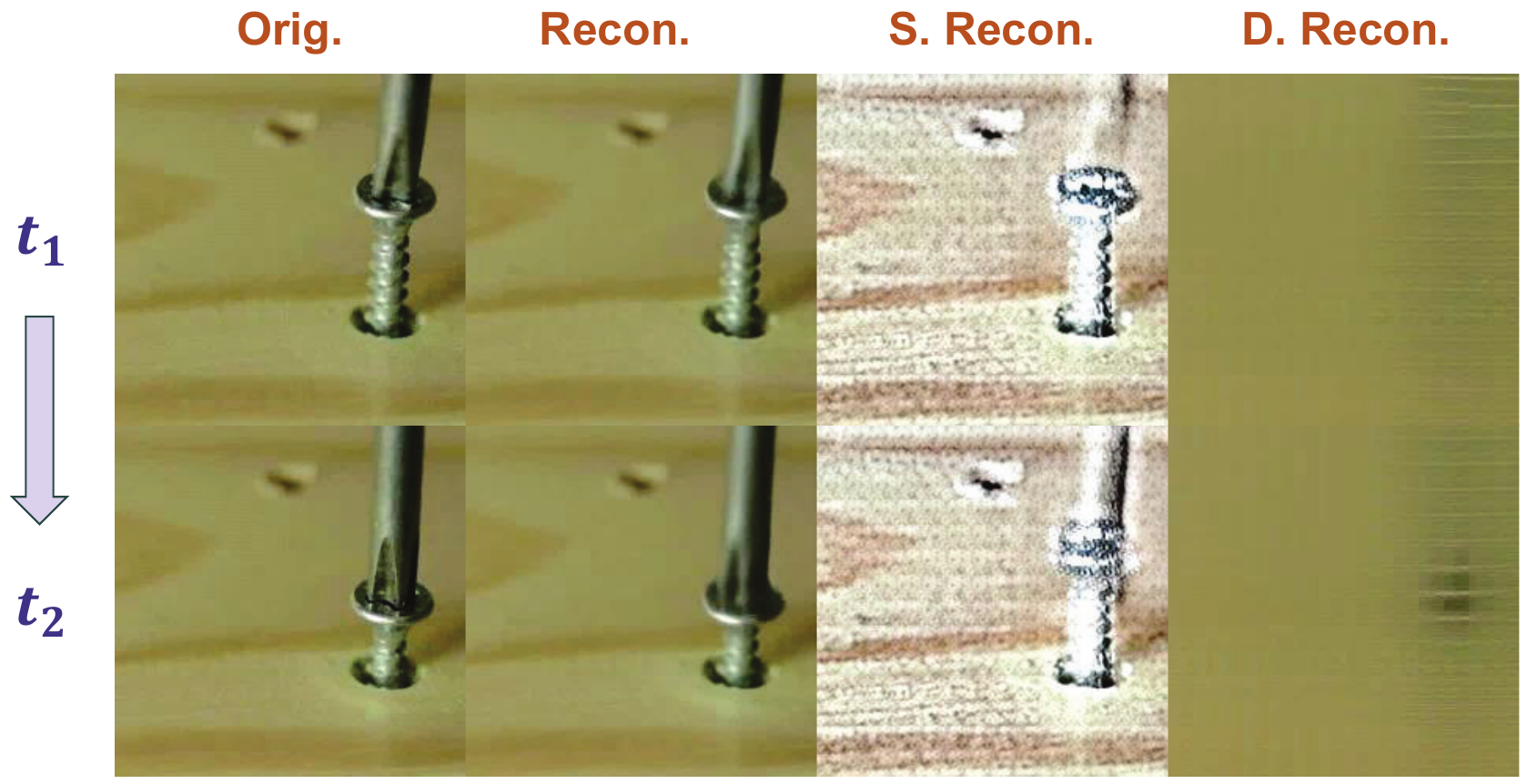}
    \caption{An example illustrating the Structure and Dynamics latents. We select two frames, $t_1$ and $t_2$, and show the original and reconstructed video frames, labeled Orig. and Recon., respectively. S. Recon. and D. Recon. refer to the reconstructed frames decoded using only the corresponding Structure or Dynamics latents. The Structure latent captures the main semantic content and overall motion trends, while the Dynamics latent encodes local details and rapid movements.
}
    \label{fig:demo}
\end{figure}

 The latent diffusion model has recently revolutionized the popular text-to-image field, with representative models such as the Stable Diffusion series~\cite{rombach2022high, podell2023sdxlimprovinglatentdiffusion,sauer2023adversarialdiffusiondistillation,esser2024scalingrectifiedflowtransformers}. In this paradigm, the image autoencoder plays a critical role by encoding the image into a compact latent space, thereby alleviating modeling complexity and improving training efficiency of the diffusion model. Recently, there has been growing interest in adapting this paradigm for video latent representation and downstream video generation tasks~\cite{sora, blattmann2023stablevideodiffusionscaling,ho2022videodiffusionmodels, cmd, blattmann2023alignlatentshighresolutionvideo,he2023latentvideodiffusionmodels}. However, due to the extra temporal consistency of videos compared to static images, simultaneously modeling visual content and temporal dependencies into a latent space presents a challenging problem~\cite{cmd}.

Upon reviewing previous works that explore the conversion of video into latent representations using autoencoders~\cite{cmd, sora, ho2022videodiffusionmodels, lavit, magvit2}, we identify two main design philosophies. First, classical approaches, represent each frame (or a group of frames) as latent vectors or tokens of uniform size~\cite{magvit, magvit2, emu,cvvae}. This method is straightforward but overlooks the redundancy between frames. Video inherently exhibits continuity, indicating that adjacent frames typically differ only slightly in details, suggesting significant potential for further compression. The second emerging approach addresses this problem by dividing the representation into two types, \ie, a single or a few content frame(s) along with several motion latent vectors~\cite{cmd,lavit,ivideogpt}. However, these decoupling methods oversimplify the dynamic nature of video content, leading to unsatisfactory generation results, such as blurred frames~\cite{cmd}.

In this paper, we propose a novel approach that encodes videos into two distinct latent spaces: \textbf{Structure Latent}, which represents the global content and movement, and \textbf{Dynamics Latent}, which captures fine-grained details and rapid motions. For instance, in the video of tightening a screw shown in \cref{fig:demo}, the main semantic content, such as the table and screw, corresponds to \textbf{Structure Latent}, while fine-grained details, such as color, texture, and rapid local movements---like the screw’s downward motion and rotation---are captured by \textbf{Dynamics Latent}. These components are combined to form the reconstructed video. To achieve this, we introduce the \textbf{\modelname} model, designed to effectively learn these interdependent latent representations. Our approach addresses the shortcomings of previous methods that often neglect dynamic content, enabling a video autoencoder capable of achieving high compression without compromising reconstruction quality.

Specifically, we utilize the Spatial-Temporal Transformer~\cite{bertasius2021spacetimeattentionneedvideo} as the backbone of our video autoencoder and introduce two submodules to extract \lf and \hf, respectively. For the former, leveraging the powerful information extraction capabilities of the Q-Former~\cite{li2023blip2bootstrappinglanguageimagepretraining} architecture, we apply it solely to the temporal dimension to extract the low-frequency changing motion trends independently of spatial location. We then further downsample the latent in the spatial dimension to remove redundant details while retaining the most important object information. For the latter, since rapid motion information can be represented in low dimensions, we first downsample the latent vectors obtained from the encoder spatially, and then we average these vectors along both the height and width dimensions to further reduce their dimensionality. Additionally, we design a mechanism to adapt the obtained latents for diffusion models by patchifying the two latent vectors and concatenating them as the training target for the diffusion model.

Through experiments, we demonstrate that our model offers several advantages: \textbf{(1) High compression rate:} The decoupling design and more compact latent representation of \modelname yield a superior compression rate, achieving around a 500$\times$ compression factor while maintaining high reconstruction quality. This significantly alleviates memory and computational burdens for downstream models, which are often challenged by the high dimensionality of video data. \textbf{(2) Effectiveness for downstream tasks:} Video autoencoders are commonly used within generative models, which require a smooth latent space. We validate this with the UCF-101~\cite{soomro2012ucf101dataset101human} dataset, where our model performs comparably to some well-established models, demonstrating its adaptability to generative tasks.   \textbf{(3) Explainability and Scalability:} As shown in~\cref{fig:demo}, we carefully design the latent space to ensure meaningful and explainable representations, and preliminary experiments also suggest that the model exhibits scalability, both of which provide opportunities for further research and improvements.

In conclusion, the main contributions of our work are summarized as follows:

\textbf{(1)} Building on the philosophy of decoupling video representation into structure and dynamics, we propose a novel video codec model, \modelname, which demonstrates effective decoupling with a compact design.

\textbf{(2)} Our \modelname achieves a high compression rate and strong reconstruction ability, and has been verified for its applicability and efficiency in generative models.

\textbf{(3)} We highlight the importance of video latent representation in current research trends and hope our \modelname can inspire or facilitate further related research.

%% file: sections/2_related.tex
\section{Related Works}
\subsection{Visual Autoencoder}
With the rapid advancement of visual generation, there has been increasing attention on visual latent representation techniques. Given that the dominant methods for generation are now diffusion models and autoregressive approaches, two primary types of corresponding representations have emerged: \textbf{(1) Continuous latent vectors:} Stable Diffusion~\cite{rombach2022high} was one of the pioneering works to utilize a Variational Autoencoder (VAE)~\cite{kingma2022autoencodingvariationalbayes} for image encoding, with a diffusion model then modeling this latent space. This approach has since inspired numerous subsequent works~\cite{podell2023sdxlimprovinglatentdiffusion, chen2023pixartalphafasttrainingdiffusion, chen2024pixartsigmaweaktostrongtrainingdiffusion, esser2024scalingrectifiedflowtransformers, wfvae,opensora,opensoraplan}. For video data, several studies have incorporated 3D convolutions~\cite{ho2022videodiffusionmodels,singer2022makeavideotexttovideogenerationtextvideo,blattmann2023alignlatentshighresolutionvideo,bartal2024lumierespacetimediffusionmodel,tang2024vidtok,cogvideox, cvvae, odvae} or spatio-temporal attention mechanisms~\cite{ho2022imagenvideohighdefinition, sora, jiang2024diveditbasedvideogeneration,Wu_2023_ICCV} into the backbone, resulting in a latent space specifically designed for video data, which facilitates more effective video generation. \textbf{(2) Discrete tokens:} In a separate line of work, influenced by the success of language modeling in the NLP community, several models have explored discrete representations of visual information. VQ-VAE~\cite{oord2018neuraldiscreterepresentationlearning} introduced a codebook into the VAE~\cite{kingma2022autoencodingvariationalbayes} training procedure to discretize the representation, while VQ-GAN~\cite{esser2021tamingtransformershighresolutionimage} incorporated adversarial training to improve the quality of generated images. Later models further refined their architectures, such as replacing CNNs with Transformers~\cite{yu2022vectorquantizedimagemodelingimproved} or improving quantization methods~\cite{mentzer2023finitescalarquantizationvqvae, magvit2}. For video data, some approaches treat frames as independent images for tokenization~\cite{gupta2022maskvitmaskedvisualpretraining,wang2022bevtbertpretrainingvideo,zhang2024video}, while others incorporate 3D architectures to capture spatio-temporal features~\cite{magvit, cmd,ge2022longvideogenerationtimeagnostic,villegas2022phenakivariablelengthvideo}. Among these, MAGVIT-v2~\cite{magvit2} has emerged as a prominent video tokenizer, proposing a look-up-free quantizer and has been widely adopted in recent models.

\begin{figure*}
    \centering
    \includegraphics[width=0.9\textwidth]{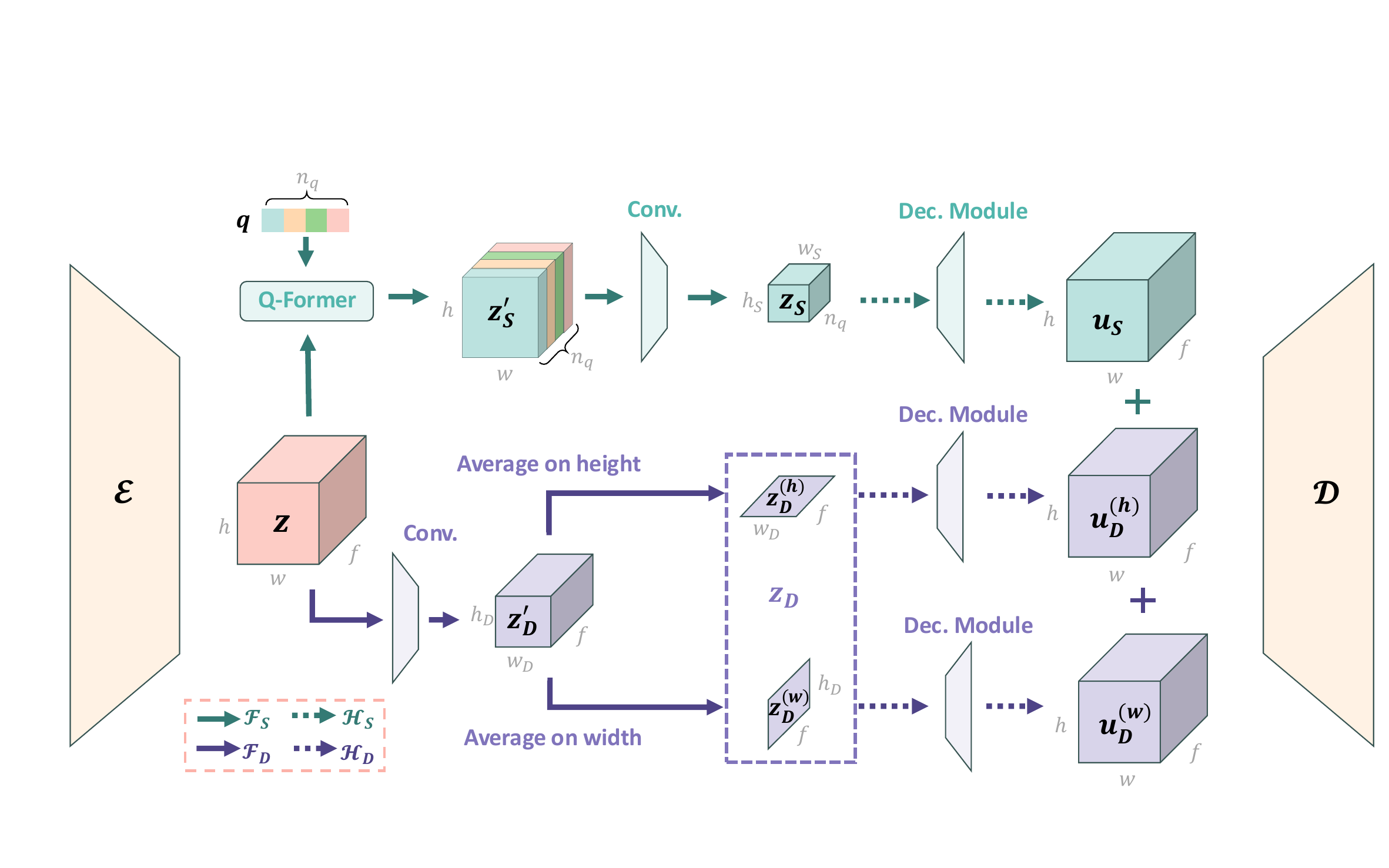}
    \caption{Details of our model. After obtaining the latent $z$ from the Encoder, the process branches into two flows. The \lf extraction module, $\mathcal{F}_{\boldsymbol{S}}$, which consists of a Q-Former and convolutional networks, extracts the \lf component $z_{\boldsymbol{S}}$. The \hf extraction module, $\mathcal{F}_{\boldsymbol{D}}$, comprising convolutional networks and an averaging operator, extracts the \hf component $z_{\boldsymbol{D}}$. Finally, using the decoding module, we align all latents to the same dimension and combine them before passing them into the Decoder.
}
    \label{fig:main}
\end{figure*}

\subsection{Video Compression and Decoupling}
Video compression is a critical challenge in computer vision, and the philosophy of decoupling has been employed in traditional video codecs for many years. For instance, MPEG-4~\cite{le1991mpeg} uses I-frames to represent key frames and macroblock motion to capture movement. Building on this concept, Video-LaViT~\cite{lavit} recently designed a pipeline that transforms key frames and motion vectors into tokens, integrating them with large language models. Other representative methods for motion representation include Motion-I2V~\cite{shi2024motioni2vconsistentcontrollableimagetovideo}, which uses pixel trajectories to capture motion, and \citep{lew2024disentangledmotionmodelingvideo}, which employs optical flow for frame interpolation. Some approaches focus on specific video types, such as GAIA series~\cite{he2024gaiazeroshottalkingavatar,wang2024instructavatartextguidedemotionmotion,yu2024makeactortalkgeneralizable} focuses on talking-face videos and uses self-cross reenactment to disentangle identity and motion, or iVideoGPT~\cite{ivideogpt}, which explores embodied videos. CMD~\cite{cmd} utilizes a weighted average of all frames to represent content, while motion is learned by a neural network. However, we identify several limitations in these methods, such as incompatibility with generative models, reliance on complex architectures, or unsatisfactory results due to excessive prior knowledge in some models. In contrast, we revisit the decoupling mechanism and propose a novel approach. Experiments demonstrate that our method has great promise, and we hope it will inspire further innovation in the community.

%% file: sections/3_method.tex
\section{Methodology}

In this section, we introduce the \modelname model. In~\cref{sec: model_overview}, we provide an overview of the architecture of \modelname. Subsequently,~\cref{sec:method_enc} describes the process of converting a video into \lf and \hf, while~\cref{sec:method_dec} delineates the process of reconstructing the video from these two latents. In~\cref{sec:train_infer}, we outline the training and inference pipelines, and lastly, in~\cref{sec:method_diff}, we discuss a design for adapting our proposed latents for use with diffusion models.

\subsection{Overall Architecture} \label{sec: model_overview}
A classical autoencoder consists of an encoder $\mathcal{E}$ and a decoder $\mathcal{D}$. Given a video $x \in \mathbb{R}^{C \times F \times H \times W}$, where $C$, $F$, $H$, 
$W$ represent the channel, number of frames, height, and width, respectively, the encoder produces a latent vector $z = \mathcal{E}(x) \in \mathbb{R}^{c \times f \times h \times w}$, where $c$, $f$, $h$, 
$w$ are corresponding dimensions with $x$ but with lower dimensions. The decoder attempts to reconstruct the input as $\hat{x} = \mathcal{D}(z) = \mathcal{D}(\mathcal{E}(x))$. The encoder and decoder are jointly trained to minimize the reconstruction loss $\mathcal{L}_{rec} = \| \hat{x} - x \|$. 

In our \modelname model, we propose decoupling a video into \lf and \hf components. As illustrated in~\cref{fig:main}, after obtaining the latent vector $z$, we introduce two processing functions, $\mathcal{F}_{\boldsymbol{S}}$ and $\mathcal{F}_{\boldsymbol{D}}$, which generate the desired latent representations $z_{\boldsymbol{S}}$ and $z_{\boldsymbol{D}}$. These procedures are described in detail in~\cref{sec:content_method} and~\cref{sec:motion_method}. For decoding, we employ two functions, $\mathcal{H}_{\boldsymbol{S}}$ and $\mathcal{H}_{\boldsymbol{D}}$, to align these latents to the same dimensional space before combining them and passing them to the decoder. The overall procedure is summarized as follows:
\begin{equation*}
\begin{aligned}
       z_{\boldsymbol{S}}, z_{\boldsymbol{D}} &= \mathcal{F}_{\boldsymbol{S}}\big(\mathcal{E}(x)\big), \mathcal{F}_{\boldsymbol{D}}\big(\mathcal{E}(x)\big) \\
    \hat{x} &= \mathcal{D}\big([\mathcal{H}_{\boldsymbol{S}}(z_{\boldsymbol{S}}); \mathcal{H}_{\boldsymbol{D}}(z_{\boldsymbol{D}})]\big)
\end{aligned}
\end{equation*}

\subsection{Encode Video into Latents} \label{sec:method_enc}

We will use Structure function and Dynamics function to extract \lf and \hf, respectively. 

\subsubsection{Structure Latent Extraction} \label{sec:content_method}

To extract the temporal low-frequency representation from the encoder's output latent $z \in \mathbb{R}^{c \times f \times h \times w}$, we employ the Q-Former, a classical interface proposed in BLIP-2~\cite{li2023blip2bootstrappinglanguageimagepretraining} that serves as a bridge between different modalities. We choose this module due to its elegant architecture and proven ability to extract semantic information from visual input. It is a Transformer~\cite{vaswani2023attentionneed} architecture with learned queries as input. In each block, the latent serves as a condition to perform cross-attention, and the last hidden states are taken as the output. In our scenario, as shown in~\cref{fig:main}, we define the query $q$ as $n_q$ tokens ($n_q \le f$) with dimension $d_q$ as input. Then, for the latent $z$, we turn it into a sequence by merging the spatial dimensions into the batch dimension, resulting in dimension $(hw, f, c)$. We then use an MLP to convert the channel dimension \( c \) into \( d_q \), and perform standard Q-Former operations along the temporal dimension. This process dynamically selects \( n_q \) representative features from the \( f \) frames. The final output \( z'_{\boldsymbol{S}} \) is obtained as:
\[
z'_{\boldsymbol{S}} = \operatorname{Qformer}(z, q) \in \mathbb{R}^{(hw) \times n_q \times d_q}
\]
Notably, when we combine the height and width dimensions into the batch dimension, it compels the Q-Former to learn the general temporal motion trends independently of location, which aligns with our expectations.

We now have the intermediate latent $z'_{\boldsymbol{S}}$, but it still faces two challenges: (1) Spatial compression has not been performed, resulting in a high product of $h$ and $w$, and (2) the dimensionality of the Q-Former's hidden state, $d_q$, remains high. To address these, we reshape $z'_{\boldsymbol{S}}$ into shape $(n_q, d_q, h, w)$ and apply several convolutional layers to downsample the spatial dimensions while using a bottleneck to reduce the channel dimension $d_q$ to a smaller size $d_{\boldsymbol{S}}$. These operations reduce the dimensionality of the final \lf while preserving main content information by eliminating detailed spatial information. Finally, we obtain the final \lf $z_{\boldsymbol{S}} \in \mathbb{R}^{n_q \times d_{\boldsymbol{S}} \times h_{\boldsymbol{S}} \times w_{\boldsymbol{S}}}$.

\subsubsection{Dynamics Latent Extraction} \label{sec:motion_method}
For dynamic local details, we consider that rapid motion information should be low-dimensional and distributed across each frame. Therefore, instead of manipulating the temporal dimension, we primarily focus on the spatial dimensions. A natural approach to reduce the dimensions is to use a spatial Q-Former to extract the most relevant spatial locations, similar to the method used for the \lf. However, this approach disrupts spatial consistency, leading to performance degradation in our experiments. 

Instead, we design an alternative approach. As shown in~\cref{fig:main}, we first downsample the latent $z$ along the spatial dimensions using convolutional layers, obtaining an intermediate result $z'_{\boldsymbol{D}}$ with dimensions $(f, c'_{\boldsymbol{D}}, h_{\boldsymbol{D}}, w_{\boldsymbol{D}})$. Inspired by~\cite{cmd}, we then average $z'_{\boldsymbol{D}}$ along the height and width dimensions to eliminate these spatial dimensions. The resulting vectors are concatenated and passed through a head $\mathcal{G}$ to reduce the channel dimension to $d_{\boldsymbol{D}}$:
\[
z_{\boldsymbol{D}} = \mathcal{G}\left([\operatorname{avg}_h(z'_{\boldsymbol{D}}); \operatorname{avg}_w(z'_{\boldsymbol{D}})]\right) \in \mathbb{R}^{f \times d_{\boldsymbol{D}} \times (w_{\boldsymbol{D}} + h_{\boldsymbol{D}})}
\]
This results in the \hf $z_{\boldsymbol{D}}$. Notably, this approach reduces the latent dimension from $\mathcal{O}(w_{\boldsymbol{D}} \cdot h_{\boldsymbol{D}})$ to $\mathcal{O}(w_{\boldsymbol{D}} + h_{\boldsymbol{D}})$, effectively extracting compact dynamic details while preserving spatial integrity.

\input{tables/rec}

\subsection{Decode Latents to Video} \label{sec:method_dec}

With the expected latents \lf and \hf obtained, we need to find a way to combine them before inputting them into the decoder. For \lf $z_{{\boldsymbol{S}}}$ with shape $(n_q, d_{\boldsymbol{S}}, h_{\boldsymbol{S}}, w_{\boldsymbol{S}})$, we apply upsampling layers to recover the spatial size and MLPs to adjust the channel dimension $d_{\boldsymbol{S}}$ and query token number $n_q$, yielding $u_{\boldsymbol{S}} \in \mathbb{R}^{c \times f \times h \times w}$. 

For \hf $z_{{\boldsymbol{D}}}$ with shape $(f, d_{\boldsymbol{D}}, w_{\boldsymbol{D}} + h_{\boldsymbol{D}})$, we process the latents for height and width separately. Specifically, for latents $z_{\boldsymbol{D}}^{(h)} \in \mathbb{R}^{f \times d_{\boldsymbol{D}} \times w_{\boldsymbol{D}}}$ and $z_{\boldsymbol{D}}^{(w)} \in \mathbb{R}^{f \times d_{\boldsymbol{D}} \times h_{\boldsymbol{D}}}$, we use MLPs $\mathcal{T}$ to recover the corresponding spatial and channel dimensions, followed by repeating along the missing spatial dimension:

\[
u_{\boldsymbol{D}}^{(h)} = \operatorname{Rep}_w (\mathcal{T}(z_{\boldsymbol{D}}^{(h)})) \in \mathbb{R}^{c \times f \times h \times w}
\]
\[
u_{\boldsymbol{D}}^{(w)} = \operatorname{Rep}_h (\mathcal{T}(z_{\boldsymbol{D}}^{(w)})) \in \mathbb{R}^{c \times f \times h \times w}
\]

Subsequently, we perform an element-wise addition of these latents and pass them to the decoder to obtain the final output video:
\[
\hat{x} = \mathcal{D}(u_{\boldsymbol{S}} + u_{\boldsymbol{D}}^{(h)} + u_{\boldsymbol{D}}^{(w)}) \in \mathbb{R}^{C \times F \times H \times W}
\]

\subsection{Training and Inference} \label{sec:train_infer}

We train all modules, including the Encoder $\mathcal{E}$, Decoder $\mathcal{D}$, latent extraction modules $\mathcal{F}_{\boldsymbol{S}}$, $\mathcal{F}_{\boldsymbol{D}}$, and decoding heads $\mathcal{H}_{\boldsymbol{S}}$, $\mathcal{H}_{\boldsymbol{D}}$, jointly to recover the input. Following the standard loss definition for image autoencoders proposed in VQ-GAN~\cite{esser2021tamingtransformershighresolutionimage}, we employ the basic reconstruction loss $\mathcal{L}_{rec}$ along with feature-level perceptual loss $\mathcal{L}_{p}$ and adversarial losses $\mathcal{L}_{GAN}$. Considering that \modelname is likely to be integrated into a generative model, we expect the latent space to be sufficiently smooth. Thus, we adopt a VAE paradigm, wherein instead of directly inputting the latents into the decoder, we introduce randomness around the latents, namely $
v(z) = \mu(z) + \sigma(z) \cdot \epsilon
$,
where $\epsilon \sim \mathcal{N}(\mathbf{0},\mathbf{I})$, and $\mu(z)$ and $\sigma(z)$ are learnable modules predicting the mean and standard deviation. To regularize this distribution, we use the KL divergence loss with the standard Gaussian distribution $
\mathcal{L}_{KL} = KL(\mathcal{N}(\mu_z, \sigma_z) || \mathcal{N}(\mathbf{0}, \mathbf{I}))$. More detailed explanations of the VAE model can be found in~\cref{sec:app_vae}. The final loss is defined as:
\[
\mathcal{L} = \mathcal{L}_{rec} + \lambda_{p}\mathcal{L}_{p} + \lambda_{GAN}\mathcal{L}_{GAN} + \lambda_{KL}\mathcal{L}_{KL}
\]

During sampling, we use the mean of the latent, \ie, $\mu(z)$. If the required latents are predicted from a generative model, we simply follow the decoding method mentioned in~\cref{sec:method_dec} to generate the final video.

\subsection{Conditional Video Generation with \modelname} \label{sec:method_diff}

Typically, \modelname is expected to connect with a generative model. Here, we present a basic design to adapt \lf and \hf for use in a diffusion model and welcome other designs from the community.

Given a video, we first apply the trained \modelname to obtain the \lf latent $z_{\boldsymbol{S}}$ and the \hf latent $z_{\boldsymbol{D}}$. The dimension of $z_{\boldsymbol{S}}$ is $(d_{\boldsymbol{S}}, n_q, h_{\boldsymbol{S}}, w_{\boldsymbol{S}})$, which resembles ``video-like'' data. For $z_{\boldsymbol{D}}$, we combine the latents along the height and width dimensions, introducing a pseudo-dimension in the second dimension to yield $(d_{\boldsymbol{D}}, 1, f, h_{\boldsymbol{D}}+w_{\boldsymbol{D}})$, effectively treating it as a single-frame video. We then apply a 3D patchification method to convert both latents into two sequences of tokens, each with dimension $d_{\textrm{Diff}}$. Since these token embeddings originate from different latents, we align them to a similar scale through normalization and then concatenate them along the length dimension to form the training target.

With the ground-truth latent training target $y_0$ and any relevant conditions $c$ (such as text or video class), we perform the standard diffusion training procedure~\cite{ddpm}. This involves sampling noise, adding it to the latent to get the noisy version $y_t$, and then attempting to remove the noise using a learnable model ${\boldsymbol{Di}}$. We utilize the current popular mechanism to predict $x_0$ directly, defined as $
\mathcal{L}_{\textrm{Diff}} = \|{\boldsymbol{Di}}(y_t, c) - y_0\|$. During sampling, we follow the DDIM~\cite{ddim} method to predict $\hat{y}_0$. We also employ classifier-free guidance~\cite{ho2022classifierfreediffusionguidance} to further enhance the model's conditioning capabilities. More details about the diffusion model can be found in~\cref{sec:app_diff}. Finally, we input the predicted latents into the decoder of \modelname to generate the final output video.

%% file: tables/rec.tex
\begin{table*}[t] 
\vspace{-0mm}
\begin{center}
\tabcolsep=0.2cm
\caption{
Quantitative comparison with baseline methods. The bold values indicate the best results, while the underlined values represent the second-best. Sem., Tempo., and Deta. refer to semantic preservation, temporal consistency, and detail retention, respectively. Our model outperforms the baselines across multiple metrics, demonstrating its superior reconstruction ability.
}
\vspace{-1mm}
	\label{tab: rec}
	\begin{tabular}{l||c|cccc|ccc}
		\toprule[1.5pt]
		 Method  &
         {Compress. Rate $\downarrow$}& {PSNR $\uparrow$} &  {LPIPS $\downarrow$} & {SSIM $\uparrow$} & 
         FVD $\downarrow$ &  {Sem. $\uparrow$} & {Tempo. $\uparrow$} & {Deta. $\uparrow$}\\        \midrule

        iVideoGPT~\cite{ivideogpt} &  $1.50\%$ & $19.353$ & $0.4677$&$0.5752$ & $1693.10$ & $4.28$ & $4.33$ & $3.59$\\
        
        MAGVIT-v2~\cite{magvit2} & $0.65\%$ & $24.351$ & $0.3347$ & $0.6877$ & $653.88$ & $4.43$ & $4.46$ & $3.97$
         \\
       % Video-LaViT (Opt) & $7.03$\%\\

          CMD~\cite{cmd}                & $6.85\%$ & $27.332$ & $0.2732$ & $\underline{0.7746}$ & $468.47$ & $4.51$ & $4.35$ & $4.22$        \\
          
        EMU-3~\cite{emu}  & $\underline{0.53\%}$ &$25.359$ &$0.2543$ & $0.7260$ & $\bf{353.71}$ & $4.69$ & $\underline{4.57}$ & $4.60$\\

        CV-VAE~\cite{cvvae}  & $\underline{0.53\%}$ &$\underline{28.063}$ &$\underline{0.2436}$ & $0.7546$ & $401.92$ & $\underline{4.70}$ & $4.51$ & $\underline{4.67}$\\

        \midrule
        \modelname~(Ours) &  $\bf{0.20\%}$ & $\bf{28.137}$ & $\bf{0.2414}$ & $\bf{0.8044}$ &$\underline{388.86}$ & $\bf{4.71}$ & $\bf{4.62}$ & $\bf{4.73}$\\      
		\bottomrule[1.5pt]
	\end{tabular}
    \vspace{-2mm}
\end{center}

\end{table*}

%% file: sections/4_exp.tex
\begin{figure*}
    \centering
    \includegraphics[width=1\textwidth]{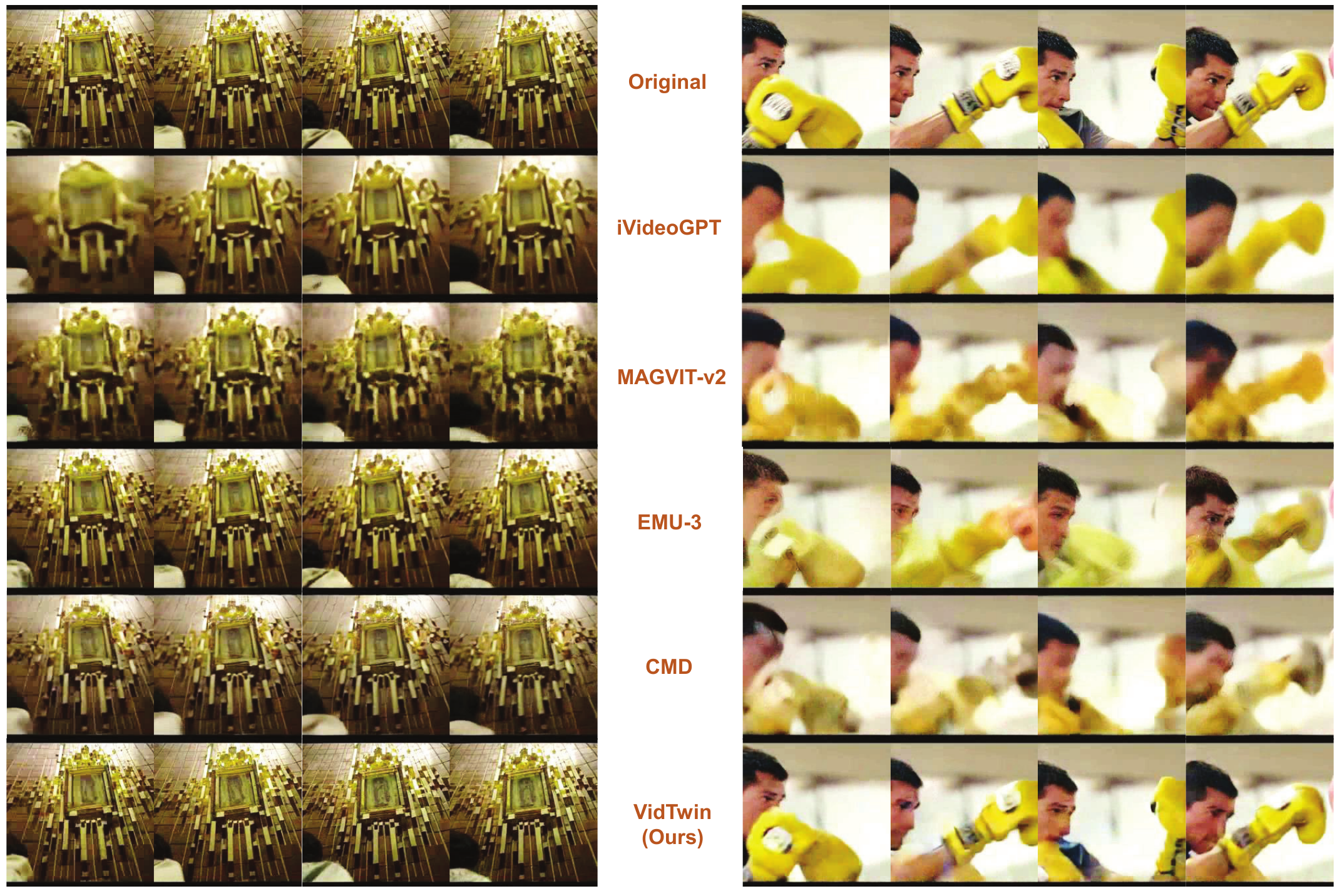}
    \caption{Qualitative comparison with baseline methods. Two examples are presented: a gradually rotating photo and a fast-motion boxing scene. \modelname demonstrates the ability to reconstruct fine details and accurately capture rapid motion.}
    \label{fig:baseline}
\end{figure*}

\section{Experiments}
We conduct experiments to validate the proposed
% advantages of our 
\modelname model, from aspects including 
% its high 
the compression rate, reconstruction ability, as well as the effectiveness and efficiency on downstream tasks.
% , and so on.

\subsection{Setup}
\subsubsection{Datasets}
% We use WebVid-10M~\cite{Bain21} as the training set for \modelname model. WebVid is 
For training, we utilize a self-collected
large-scale text-video dataset, containing 10 million video-text pairs.
% scraped from the website. 
Considering the broad variety of content and motion speed in this dataset, we believe training on this dataset is a good choice to fulfill our design philosophy. For evaluation, we use the MCL-JCV dataset~\cite{wang2016mcl}, which is a classical dataset for evaluating video compression quality. Moreover, to verify the adaptability of the latent emitted by our model to generative models, we evaluate the class-conditioned video generation ability on the UCF-101~\cite{soomro2012ucf101dataset101human} dataset, which provides 101 different classes of motion videos.

\subsubsection{Implementation Details}
We train our model on 8 fps, 16-frame, 224 $\times$ 224 video clips and evaluate on 25 fps, 16-frame, 224 $\times$ 224 video clips. The backbone of our model is a Spatial-Temporal Transformer~\cite{bertasius2021spacetimeattentionneedvideo} with a hidden dimension of 768 and a patch size of 16. Both the encoder and decoder consist of 16 layers, each with 12 attention heads, resulting in a total of about 300M parameters. For the latent representation, we configure one setting with $h_{\boldsymbol{S}} = h_{\boldsymbol{D}} = 7$, with dimensions 4 and 8, respectively. This results in two latent representations of sizes $7 \times 7 \times 16 \times 4$ and $16 \times 14 \times 8$ for a $16 \times 3 \times 224 \times 224$ video clip. Unless otherwise specified, all results in this paper use this configuration. The model is trained on 4 $\times$ A100 GPUs. We use the Adam optimizer with a learning rate of 1.6e-4. Additional hyperparameters and model settings are provided in~\cref{sec:app_imp}.

\subsection{Baselines and Evaluation Metrics}\label{sec:exp_baseline}

We select several state-of-the-art baselines, including models that represent videos as latents with uniform size, such as MAGVIT-v2~\cite{magvit2}, CV-VAE~\cite{cvvae} and the visual tokenizer of EMU-3~\cite{emu}, as well as models that decouple content and motion, like CMD~\cite{cmd} and iVideoGPT~\cite{ivideogpt}. Additional analysis of other concurrent baselines is provided in~\cref{sec:app_baselines}. We compare these baselines with our model using standard reconstruction metrics, including PSNR~\cite{psnr}, SSIM~\cite{wang2004image}, LPIPS~\cite{zhang2018unreasonableeffectivenessdeepfeatures}, and FVD~\cite{fvd}. Additionally, we conducted a human evaluation by inviting 15 professional evaluators to assess the results based on three criteria: semantic preservation (Sem.), temporal consistency (Tempo.), and detail retention (Deta.). Each evaluator is presented with 20 samples and asked to rate each on a scale from 1 to 5. The final evaluation score is computed as the Mean Opinion Score (MOS), representing the average rating across evaluators. Moreover, we also introduce the Compression Rate~(Compress. Rate) metrics, which we define as the ratio between the dimension of the latent space (or token embeddings) used in the downstream generative model, and the input video's dimension.

\subsection{Reconstruction Quality}\label{sec:exp_recon}

As shown in \cref{tab: rec}, our model achieves state-of-the-art performance across most objective and subjective metrics, demonstrating strong capabilities in video reconstruction. The vision tokenizer of EMU-3 achieves the best FVD score and good reconstruction ability, likely due to the large dataset it was trained on (i.e., InternVid~\cite{wang2024internvidlargescalevideotextdataset}). While most models perform well in semantic preservation and temporal consistency, they vary significantly in detail retention, where our model outperforms the others. Additionally, our model utilizes a highly compact latent space, approximately 2.5 to 30 times smaller than those of the baselines. It is encouraging to see that our model achieves comparable or superior reconstruction quality with such a low-dimensional latent space, highlighting the efficiency and effectiveness of \modelname. We also train our architecture at different parameter scales, and larger models perform even better. This scalability is likely due to our Transformer-based architecture; further details can be found in~\cref{sec:app_scale}.

Furthermore, we conduct case studies, with qualitative results shown in~\cref{fig:baseline} (zoom in to observe finer details). For the left case, we observe that our model effectively captures local details and the gradual rotation of the object. In comparison, baselines such as CMD show blurred edges and incomplete rotation. In the right case, featuring the fast motion of a man boxing, all baselines struggle to accurately capture the rapid movements, resulting in ghosting artifacts. In contrast, \modelname produces significantly clearer results, demonstrating the effectiveness of our decoupling strategy for capturing both low-frequency changing objects and rapid local motion. More cases are presented in~\cref{sec:app_rec}.
\begin{figure}
    \centering
    \includegraphics[width=1\linewidth]{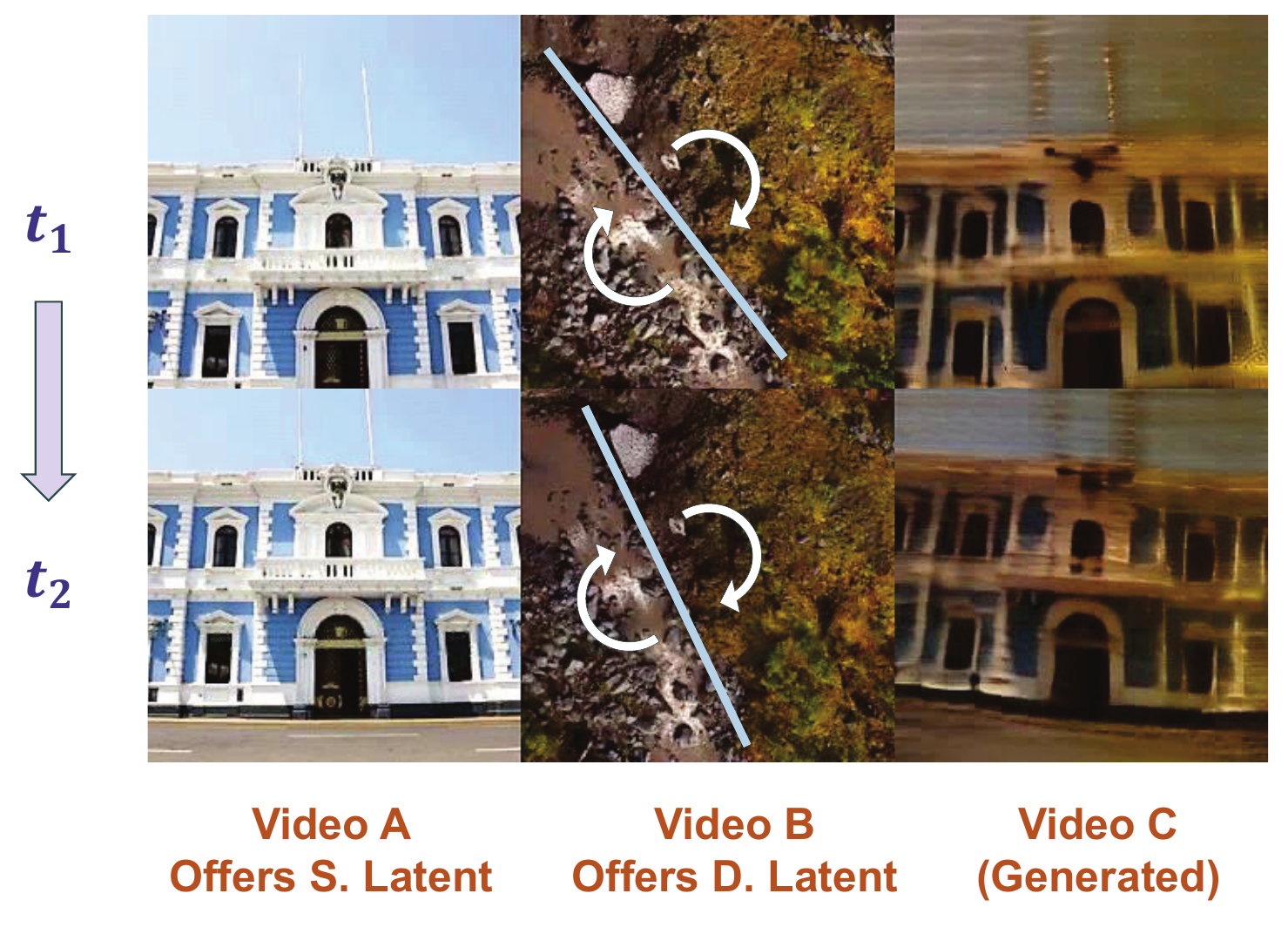}
    \caption{An illustration of a cross-replacement example, where Video C is generated using the \lf from Video A and the \hf from Video B.
        }
    \label{fig:cross}
\end{figure}

\subsection{Further Analysis}

As highlighted in~\cref{sec:intro}, our \modelname not only demonstrates strong reconstruction capabilities but also excels in explainability, efficiency, and adaptability with generative models. In this section, we provide evidence to support these claims.

\subsubsection{Explorations on the Roles of Latents} \label{sec: exp_role}
In \modelname, we design two distinct latents: \lf for the main object and overall movement trend, and \hf, which captures local details and rapid
motions. We present two experiments that provide insight into their respective roles.

First, as discussed in \cref{sec:method_dec}, we perform element-wise addition of the latents before inputting them into the decoder. This setup enables us to explore the outputs generated when each latent is passed through the decoder individually, \ie, generating results from $\mathcal{D}(u_{\boldsymbol{S}})$ and $\mathcal{D}(u_{\boldsymbol{D}})$. An example provided in \cref{fig:demo} of the~\cref{sec:intro} illustrates the distinct differences between the two latents using a scenario involving the screwing process. As observed, the \lf captures the main semantic content, such as the table and screw, while the \hf captures fine-grained details, including color and rapid local movements of the screw. Notably, in frame $t_2$, where the screw drops, the video generated by the \lf shows only a slight change, whereas the one generated by the \hf captures this immediate movement. This demonstrates the distinction between low-frequency and high-frequency movement trends.

\begin{figure}
    \centering
    \includegraphics[width=0.9\linewidth]{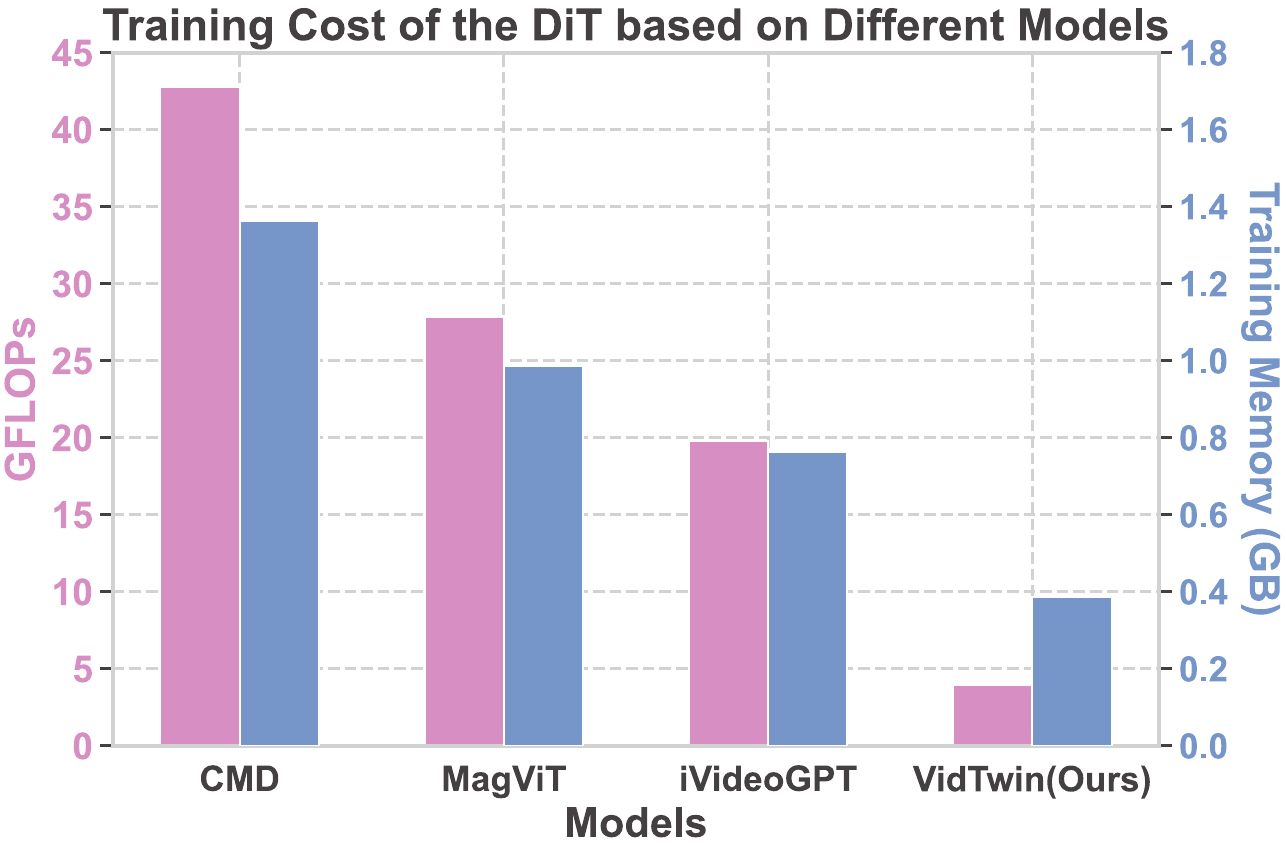}
    \caption{We present the FLOPs and training memory costs of the unified generative model, as applied to our model and the baselines.
}
    \label{fig:compute}
\end{figure}

Second, we conduct a cross-reenactment experiment in which we combine the \lf from one video, $A$, with the \hf from another video, $B$, to observe the generated output from the decoder, i.e., generating $\mathcal{D}(u^A_{\boldsymbol{S}}, u^B_{\boldsymbol{D}})$. As shown in~\cref{fig:cross}, the generated video inherits the main object (house) and overall structure from Video $A$, which provides \lf, while the local color comes from Video $B$, which provides \hf. Notably, we observe that the movement in the generated video inherits the rapid rotation from Video $B$, while adjusting the gradually downward camera view according to the scene in Video $A$. This further validates our motivation to decouple video content into overall structure and detailed dynamics. We provide additional examples for both settings in~\cref{sec:app_moreexp}.

One additional note is that, as suggested by the name \modelname, the \lf and \hf latents work together to generate the final video. These separate analyses are intended to offer a glimpse into the roles of each latent, but it is important to note that isolating them inevitably introduces information loss. In future work, we plan to explore additional methods for better understanding the intrinsic information stored in these separate latents.

\subsubsection{Computation Resource Analysis for Generative Models} \label{sec:exp_compute}
Through our decoupling design, we reduce redundancy, resulting in compact latents with a high compression rate. A key advantage of having lower-dimensional latents is the reduced computational resource requirements for downstream tasks. To demonstrate this, we compare the FLOPs and memory consumption of generative models based on representative baselines. For a fair comparison, instead of using the original generative models from their respective papers, which vary significantly, we construct a pseudo uniform DiT~\cite{peebles2023scalablediffusionmodelstransformers} architecture with a uniform patch size, focusing solely on resource consumption rather than generative ability. The results are shown in~\cref{fig:compute}. As observed, the downstream diffusion model that fits our latent space, which has a higher compression rate, requires significantly fewer FLOPs and less training memory (4 to 8 times and 2 to 3 times smaller than the baselines, respectively). This reduction in resource consumption leads to improved deployment efficiency. Furthermore, given the smaller dimension of our latent space, it is possible to use a smaller diffusion model to fit the distribution, further reducing resource requirements. Additional details about the pseudo DiT model used can be found in~\cref{sec:app_dit}.

% flops; memory; compression rate; some bar charts
\subsubsection{Generative Quality of Diffusion Models} \label{sec:exp_diff}
\input{tables/gen}
As shown in \cref{sec:method_diff}, we design a basic method to adapt our latent representations to the generation framework of a DiT-based diffusion model. We evaluate the proposed method on the UCF-101 dataset~\cite{soomro2012ucf101dataset101human} for class-conditional video generation, with the results reported in \cref{tab:gen}. Our model achieves performance comparable to several existing methods. It is important to note that the main focus of this paper is not on generation, and we have implemented only a simple baseline model to evaluate the adaptability of our approach to the diffusion framework. Despite this, the results are promising and demonstrate that the latent space in \modelname is well-suited for downstream generative tasks. We believe that with a more refined design, a larger dataset, and the incorporation of additional techniques during training, a generation model based on our latent space will achieve even better performance.

\subsection{Ablation Studies}

We conduct an ablation study to assess the impact of our proposed designs by removing each one. The experiments are evaluated using the same number of training steps, and the results are presented in~\cref{tab:ablation}.
The findings can be summarized as follows: \textbf{(a)} When we omit the disentangling paradigm and use a single latent with a similar compression rate, performance drops significantly, demonstrating that our decoupling approach not only produces meaningful latent representations but also enhances performance at the same compression rate. \textbf{(b)} As discussed in~\cref{sec:motion_method}, replacing the averaging method with a Spatial Q-Former to further compress the spatial dimensions of \hf results in poorer performance, likely due to the disruption of spatial arrangement. \textbf{(c)} We propose using a Q-Former to extract \lf. When we replace it with simple convolution layers and an MLP to decrease the temporal dimension, performance degrades, highlighting the superior semantic extraction capability of the Q-Former. \textbf{(d)} As mentioned in~\cref{sec:content_method}, moving the spatial dimensions into the batch dimension to obtain location-independent latents is crucial. Without this, and by placing them into the hidden states dimension instead, we observe a noticeable performance loss.

\input{tables/ablation}

%% file: tables/gen.tex
\begin{table}[t]
\begin{center}
\small
\tabcolsep=0.1cm
% \footnotesize
\caption{The generative ability of our model and the baselines, as tested on UCF-101.
}
\vspace{-2mm}
	\label{tab:gen}
	\begin{tabular}{l|cccc}
		\toprule[1.5pt]
	    Models & TATS~\cite{ge2022longvideogenerationtimeagnostic} & MAGVIT-v2~\cite{magvit2} & Video-LaViT~\cite{lavit} & Ours \\
		\midrule
            FVD $\downarrow$ & $332$ & $58$  & $275$ & $193$  \\
       
		\bottomrule[1.5pt]
	\end{tabular}
\end{center}
\vspace{-2mm}
\end{table}

%% file: tables/ablation.tex
\begin{table}[t]
\begin{center}
\small
\caption{Ablation studies on the proposed
techniques.}
\vspace{-2mm}
	\label{tab:ablation}
	\begin{tabular}{l|cc}
		\toprule[1.5pt]
	    Methods & PSNR$\uparrow$ & SSIM$\uparrow$  \\
		\midrule
            \modelname  & $\bf{26.116}$ & $\bf{0.731}$    \\
            \midrule
        (a) w/o Disentanglement   & $23.512$ & $0.654$  \\
        (b) w/o D. Latent Avg.   & $24.835$ & $0.693$  \\
        (c) w/o S. Latent Qformer   & $25.386$ & $0.702$  \\
        (d) w/o S. Latent Move Spa. & $23.169$ & $0.630$\\
		\bottomrule[1.5pt]
	\end{tabular}
\end{center}
\vspace{-2mm}
\end{table}

%% file: sections/5_conclusion.tex
\section{Conclusion}
In this paper, we present \modelname, a novel model for video latent representation. \modelname incorporates carefully designed submodules within an Encoder-Decoder framework to effectively separate Structure and Dynamics latent spaces. Through extensive experiments, we demonstrate that \modelname achieves high compression rates, has a simple architecture, and performs well in downstream generative tasks. Additionally, inspired by~\cite{wu2024janusdecouplingvisualencoding}, the \lf space in our model appears well-suited for visual understanding tasks, which we plan to explore in future work. Finally, our approach provides explainability and scalability, making it valuable for future research. We hope that our work will inspire new decoupling techniques in the video community and contribute to advancements in both video generation and broader multimodal applications.

\section*{Acknowledgments}
This work was conducted during an internship at Microsoft Research Asia and was supported by the company. Additionally, it was partially supported by the National Natural Science Foundation of China under Grant No. 92470205.

%% file: sections/appendix.tex
\clearpage
\appendix
% \setcounter{page}{1}
% \maketitlesupplementary

% \tableofcontents

\section{Additional Experimental Results} \label{sec:app_moreexp}

To enhance the visual experience, we strongly encourage viewing the videos \href{https://vidtwin.github.io/}{on the website}.

\subsection{Additional Reconstruction Examples} \label{sec:app_rec}
\input{tables/scale}

\begin{figure*}[ht]
    \centering
    \includegraphics[width=\linewidth]{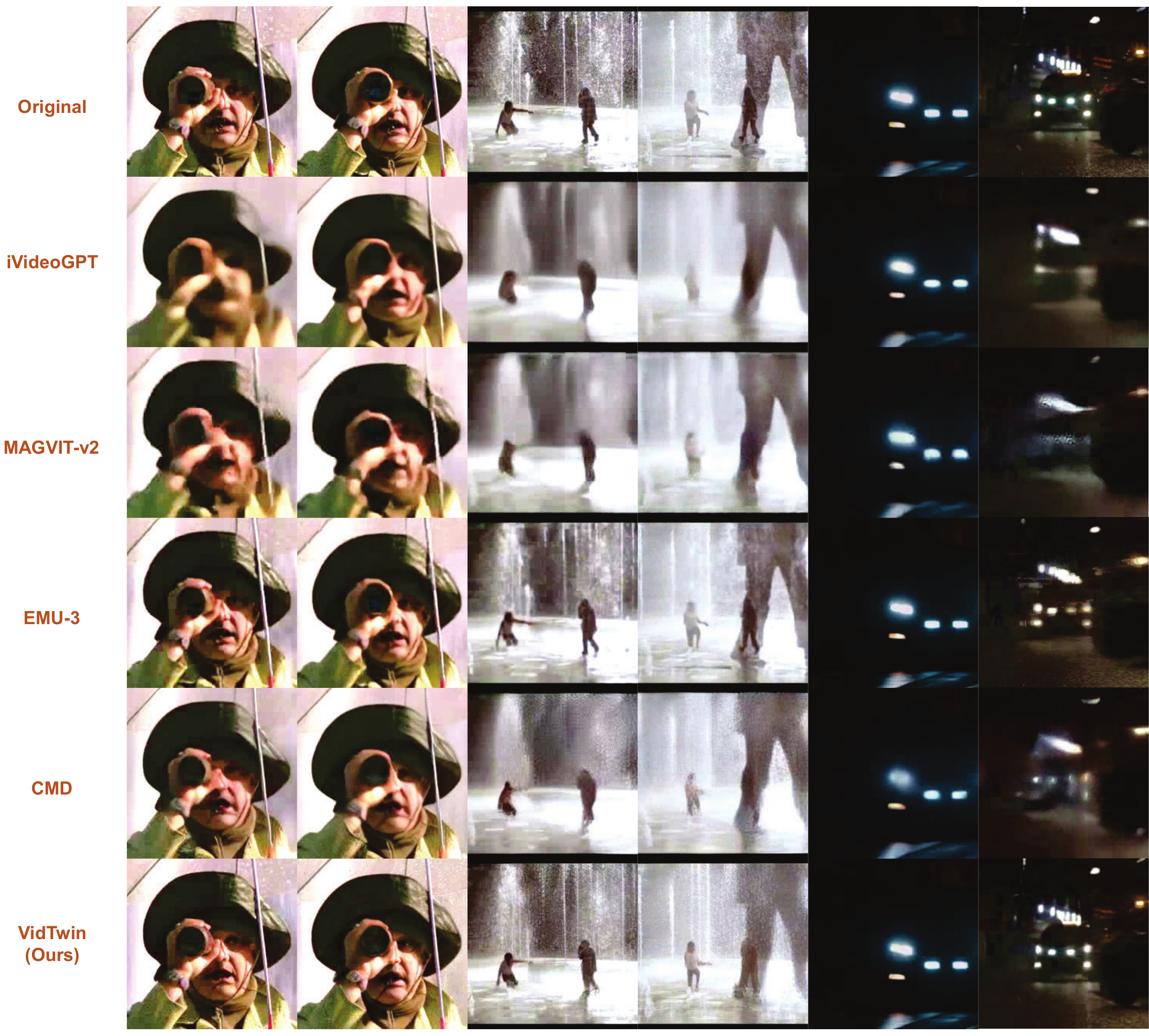}
    \caption{Additional reconstruction cases comparing our \modelname model with baselines. Zoom in to observe finer details.}
    \label{fig:app_recon}
\end{figure*}

\cref{fig:app_recon} presents additional reconstruction examples. By zooming in, one can observe that our \modelname effectively captures intricate details, such as raindrops in the first and second cases. Moreover, by decoupling structural and dynamic motion features, our model excels at preserving rapid motion dynamics. For example, in the third case, \modelname accurately reproduces the light trails of a fast-moving car, where other baselines fail to do so.

\subsection{Additional Decoupling Examples}
\begin{figure*}[ht]
    \centering
    \includegraphics[width=\linewidth]{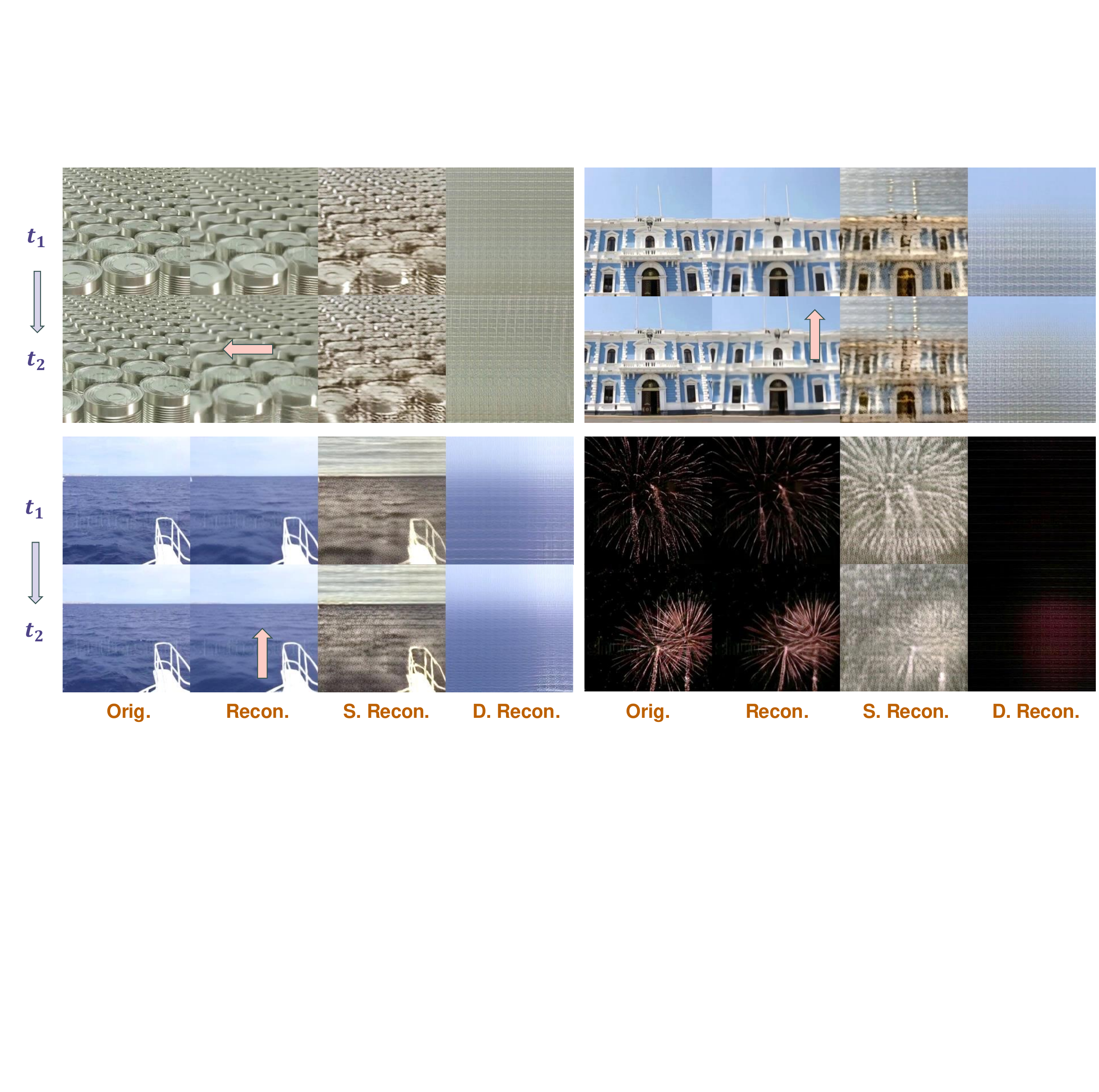}
    \caption{Additional examples of decoupling \lf and \hf.}
    \label{fig:app_disen}
\end{figure*}

In~\cref{sec: exp_role}, we demonstrated the ability to separately recover the \lf and \hf components. Additional examples are shown in Figure~\ref{fig:app_disen}. Videos generated using \lf predominantly capture primary structures and main objects, while those generated with \hf focus on colors and rapid movements.

A notable example is observed in the bottom-right case, where fireworks visible in the first frame disappear in the second. However, the \lf-generated video retains the fireworks from the first frame, demonstrating that \lf effectively encodes low-frequency, gradually evolving information.

We would like to emphasize that our primary objective is not to completely decouple structure and dynamics, as this is a challenging problem even for humans. Instead, we observe potential in reducing temporal redundancy in video representation. Based on this observation, we designed an algorithm that strives to decouple video content into these two spaces. Therefore, the cross-reenactment experiment was only designed to intuitively demonstrate the roles of the two latents rather than being specifically optimized for cross-reenactment videos.

\subsection{Additional Cross-Reenactment Examples}

\begin{figure*}[ht]
    \centering
    \includegraphics[width=\linewidth]{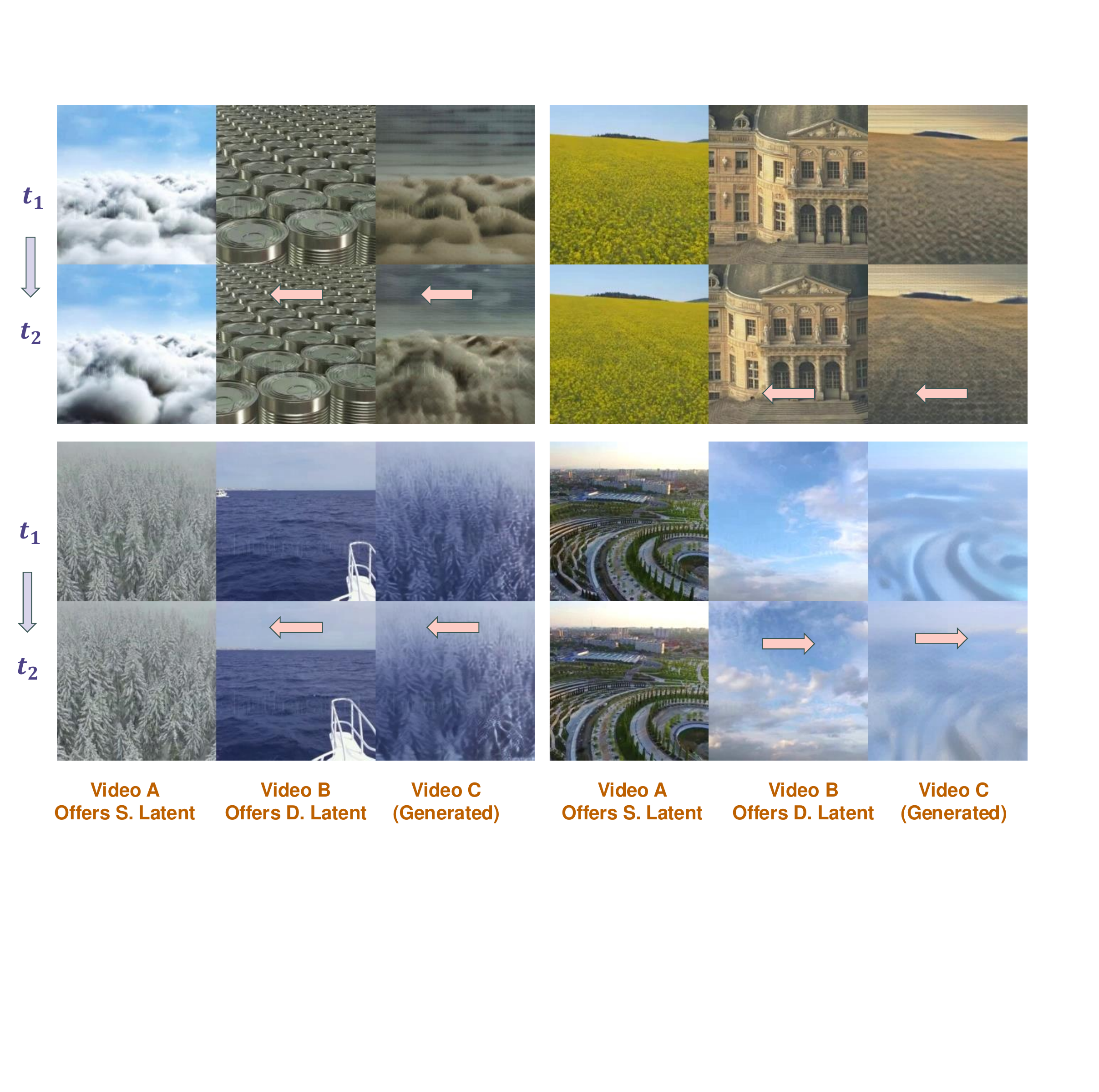}
    \caption{Additional examples of cross-reenactment.}
    \label{fig:app_cross}
\end{figure*}

\cref{fig:app_cross} provides further examples of the cross-reenactment experiments described in~\cref{sec: exp_role}. In these examples, the generated videos inherit the basic structure from Video $A$ while incorporating local details and motions from Video $B$. Notably, motion patterns such as horizontal movements and wave-like motions, as seen in the two bottom cases, are effectively transferred.

\input{tables/baselines_app}
\subsection{Comparison with Concurrent Baselines}\label{sec:app_baselines}

Recent works have explored the field of video autoencoders~\cite{wfvae,odvae,opensora,opensoraplan,cogvideox}. We observe that most of these baselines still fall into the category of methods that represent frames as latent vectors of uniform size, as discussed in~\cref{sec:intro}. A comparison between our model and these baselines is presented in~\cref{tab:app_baselines}.  
Notably, our model achieves performance comparable to state-of-the-art methods. CogVideoX~\cite{cogvideox} demonstrates impressive results, likely due to its large-scale training data. Additionally, even our highest compression rate model achieves a lower compression rate than other models (typically 0.6\% with $8,8,4$).

\input{tables/ratio}

\section{Additional Analysis for VidTwin}
\subsection{Definition of Compression Rate and the Trade-off with Reconstruction Quality.}
Differs from the typical representation that uses the downsampling factors for height, width, and number of frames for compression rate, we define it as the ratio between the dimension of the latent used in the downstream model and the input video’s dimension. For example, the typical downsampling factor ($8,8,4$) with channels 4 corresponds to a compression rate of $0.65\%$ in our definition. Additionally, we present the trade-off between compression rate and reconstruction quality in~\cref{tab:compr_ration}. As shown, lower compression rates generally result in better reconstruction quality. 
\subsection{Initial Scalability Exploration} \label{sec:app_scale}
In~\cref{sec:exp_recon}, we described training our architecture at varying parameter scales and observed consistent performance improvements with larger models. \cref{tab:scale} summarizes the configurations of each model, evaluated at the same training step. The results demonstrate a steady enhancement in reconstruction quality with increasing model size. In future work, we plan to explore additional model scales and investigate potential scaling laws, including exponential trends and other patterns.

\input{tables/other}
\subsection{Performance of VidTwin with Increased Frames and Higher FPS}  
We selected the same subjective evaluation subset as in~\cref{sec:exp_baseline} and sampled videos with 32 frames and 40 FPS. A new user study was conducted, and the results are presented in~\cref{tab:other}. The findings indicate that VidTwin maintains strong performance with an increased number of frames and a higher frame rate.  

\subsection{Failure modes of VidTwin}
\begin{figure}
    \centering
    \includegraphics[width=\linewidth]{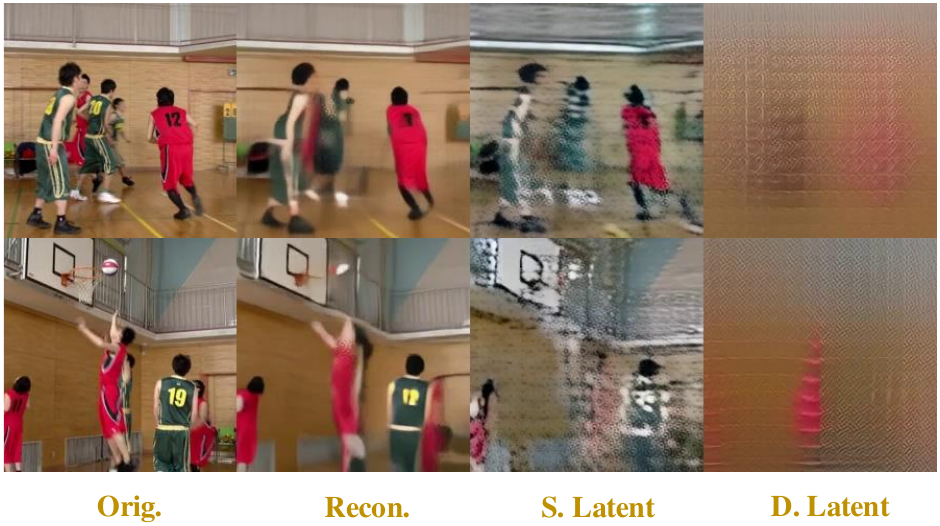}
    \caption{Failure modes of VidTwin.}
    \label{fig:app_failure}
\end{figure}

We provide a failure case in~\cref{fig:app_failure}, depicting a basketball scene with fast player movements and camera motion. While the background remains well-preserved, the fast-moving individuals appear blurred. In terms of components, the S. Latent captures the background but becomes blurred for the players, which is expected as it encodes slowly changing semantic information. The D. Latent captures the fast-changing players but struggles to accurately integrate them into the reconstructed video due to their extremely rapid movement. We plan to address this issue by pretraining on low-fps videos and fine-tuning on high-fps videos in future work.

\section{Additional Information on Experimental Settings}
\subsection{Baselines and Compression Rates}
This section provides details on the baselines used in our evaluation and discusses their compression rates, as outlined in~\cref{sec:exp_baseline}. Notably, MAGVIT-v2~\cite{magvit2}, iVideoGPT~\cite{ivideogpt}, and CMD~\cite{cmd} do not offer official code or pretrained checkpoints. Therefore, we reimplement these methods based on the descriptions provided in their respective papers.

\paragraph{MAGVIT-v2~\cite{magvit2}:} 
MAGVIT-v2 employs 3D causal CNN layers to downsample videos into latents, with a temporal downsampling factor of 4 and spatial downsampling factor of 8. The latent dimension is set to 5, as reported in the paper, resulting in a compression rate of: 
\[
\frac{5}{3 \times 4 \times 8 \times 8} \approx 0.65\%.
\]

\paragraph{EMU-3~\cite{emu}:} 
EMU-3 is a generative model proposed by BAAI\footnote{\url{https://www.baai.ac.cn/}}. For our evaluation, we primarily utilize its video tokenizer, which is based on SBER-MoVQGAN\footnote{\url{https://github.com/ai-forever/MoVQGAN}}. This tokenizer incorporates two temporal residual layers with 3D convolutional kernels in both the encoder and decoder modules, enhancing video tokenization. Similar to MAGVIT-v2, it achieves a 4$\times$ temporal compression and 8$\times$8 spatial compression. The compression rate, with a latent size of 4, is calculated in the same manner.

\paragraph{CV-VAE~\cite{cvvae}}

CV-VAE is a video VAE of latent video models, designed to have a latent space compatible with that of a given image VAE, such as the image VAE in Stable Diffusion (SD). In terms of compression rate, it matches that of EMU-3, achieving $4\times$ temporal compression and $8\times8$ spatial compression.

\paragraph{CMD~\cite{cmd}:} 
CMD decouples video representations into content frames and motion latents. For a video of size $(c, f, h, w)$, the content frame has dimensions $(c, h, w)$, and the motion latent is $(d, h + w, f)$, where \(d\) is the dimension of the motion vector. Based on the settings described in the paper, the compression rate is:
\[
\frac{1}{f} + \frac{d(h+w)}{chw} = \frac{1}{16} + \frac{2 \times 224 \times 32}{3 \times 224 \times 224} \approx 6.9\%.
\]
The primary bottleneck lies in the content frame, and we hypothesize that longer video clips could reduce the compression rate (though at the potential cost of performance).

\paragraph{iVideoGPT~\cite{ivideogpt}:} 
iVideoGPT employs a conditional VQGAN~\cite{esser2021tamingtransformershighresolutionimage} with dual encoders and decoders. The context frames \(1:T_0\) are encoded using \(N_0\) tokens, while subsequent frames are encoded with fewer tokens (\(n\)), conditioned on the context tokens to capture the essential dynamics. The compression rate is given by:
\[
\frac{N_0 d + n(T-T_0)d}{C \times T \times H \times W},
\]
and, based on the information in the paper, we calculate it as:
\[
\frac{2 \times 16^2 \times 64 + 14 \times 4^2 \times 64}{3 \times 16 \times 256^2} \approx 1.5\%.
\]

\subsection{Pseudo DiT for Resource Consumption Evaluation} \label{sec:app_dit}
Our \modelname model offers a highly compressed latent space, significantly reducing the resource requirements of downstream generative models. To validate this, in~\cref{sec:exp_compute}, we compare the performance of a generative model applied to the latent spaces produced by \modelname and the baselines.

For a fair comparison, we utilize the same DiT~\cite{peebles2023scalablediffusionmodelstransformers} architecture in all experiments. The configuration includes 6 layers, 8 attention heads, a hidden dimension size of 512, and a feed-forward network (FFN) dimension of 2048, resulting in a total of 12,610,560 parameters. Additionally, a unified patch size of 2 is used for all dimensions.

We calculate the FLOPs using a single sample (batch size $=$ 1). For memory consumption, we employ the Adam~\cite{kingma2017adammethodstochasticoptimization} optimizer and record the maximum GPU memory usage during training.

\section{Implementation Details} \label{sec:app_imp}
\subsection{Model Details}
As described in~\cref{sec: model_overview}, our \modelname adopts an Encoder-Decoder architecture. Specifically, we utilize a Spatial-Temporal Transformer~\cite{bertasius2021spacetimeattentionneedvideo} backbone. In each block, spatial attention is first applied to the height and width dimensions, followed by temporal attention along the temporal dimension. Temporal attention uses causal masking, ensuring that earlier frames do not attend to later ones, similar to the configuration in MAGVIT-v2~\cite{magvit2}. We evaluate three different scales (outlined in~\cref{tab:scale}) by adjusting the depth, hidden state dimensions, and other parameters. For spatial dimensions, a patch size of 16 is used for both height and width, while for the temporal dimension, the patch size is set to 1.

The Q-Former~\cite{li2023blip2bootstrappinglanguageimagepretraining}, employed for extracting \lf components, consists of 6 layers with a hidden dimension of 64 and 8 attention heads. For downsampling, we primarily use convolutional layers with a stride of 2, while upsampling is performed using Upsample layers with a factor of 2. By varying the number of convolutional layers, latents of different sizes can be generated.

\input{tables/settings}
Recommended latent size settings are as~\cref{tab:latent_sizes}.
From our experiments, we see that these configurations exhibit minimal performance differences, allowing users to select a setting based on specific requirements.

\subsection{Data and Training Details}
The key hyperparameters for training data and optimization are summarized as~\cref{tab:training_config}.
\input{tables/hyper}

\subsection{Diffusion Model Details} \label{sec:app_diff}
In~\cref{sec:exp_diff}, we describe the design of a diffusion model tailored to the latent space of our \modelname model. This model adopts the DiT~\cite{peebles2023scalablediffusionmodelstransformers} architecture with 18 layers and a hidden state size of 1152. Conditioning is introduced via cross-attention, and for the UCF-101 dataset~\cite{soomro2012ucf101dataset101human}, we use a 256-dimensional vector to encode the class information.

The diffusion process consists of 1000 steps, with DDIM~\cite{ddim} used as the sampling strategy and 50 steps for inference. Classifier-free guidance~\cite{ho2022classifierfreediffusionguidance} is applied, where conditioning is randomly dropped in 20\% of the samples during training. The classifier-free guidance weight is set to 5 during sampling.

For training, we use the Adam optimizer~\cite{kingma2017adammethodstochasticoptimization} with $\beta_1 = 0.9, \beta_2 = 0.999$. The learning rate is managed with a Lambda scheduler and includes 10,000 warmup steps. Training is conducted on 8 $\times$ 40G A100 GPUs, with an input configuration of 16 video frames at a resolution of 224.

\section{Basics for Diffusion Models and VAE}
\subsection{Basics for Diffusion Models}

Diffusion models are a class of emerging generative models designed to approximate data distributions. The training process consists of two phases: the forward diffusion process and the backward denoising process. Given a data point sampled from the real data distribution, $x_0 \sim q(x)$\footnote{We follow the notation and derivation process of \url{https://lilianweng.github.io/posts/2021-07-11-diffusion-models.}}, the forward diffusion process gradually adds Gaussian noise to the sample, generating a sequence of noisy samples $x_1, \dots, x_T$. The noise scales are controlled by a variance schedule $\beta_t \in (0, 1)$, and the density can be expressed as:
\[
q(x_t|x_{t-1}) = \mathcal{N}(x_t; \sqrt{1-\beta_t}x_{t-1}, \beta_t\mathbf{I}).
\]

Using the reparameterization trick~\citep{ddpm}, this process allows for sampling at any arbitrary time step in closed form:
\begin{equation*} \label{equ:diffuse}
q(x_t|x_0) = \mathcal{N}(x_t; \sqrt{\overline{\alpha}_t}x_0, \sqrt{1-\overline{\alpha}_t}\mathbf{I}),
\end{equation*}
where $\alpha_t = 1 - \beta_t$ and $\overline{\alpha}_t = \prod_{i=1}^t \alpha_i$. 
From this, it is evident that as $T \to \infty$, $x_T$ converges to an isotropic Gaussian distribution, aligning with the initial condition used during inference.

However, obtaining a closed form for the reverse process $q(x_{t-1}|x_t)$ is challenging. When $\beta_t$ is sufficiently small, the posterior also approximates a Gaussian distribution. In this case, a model $p_{\theta}(x_{t-1}|x_t)$ can be trained to approximate these conditional probabilities:
\[
p_{\theta}(x_{t-1}|x_t) = \mathcal{N}(x_{t-1}; \mu_{\theta}(x_t, t), \Sigma_{\theta}(x_t, t)),
\]
where $\mu_{\theta}(x_t, t)$ and $\Sigma_{\theta}(x_t, t)$ are parameterized by a denoising network $f_{\theta}$, such as a U-Net~\cite{ronneberger2015unetconvolutionalnetworksbiomedical} or a Transformer~\cite{vaswani2023attentionneed}. By deriving the variational lower bound to optimize the negative log-likelihood of $x_0$, \citet{ddpm} introduces a simplified DDPM learning objective:
\[
\mathcal{L}_{\text{simple}} = \sum_{t=1}^T \mathbb{E}_q \big[\|\epsilon_t(x_t, x_0) - \epsilon_{\theta}(x_t, t)\|^2 \big],
\]
where $\epsilon_t$ represents the noise added to the original data $x_0$. In our work, we adopt a simpler architecture that directly predicts $x_0$, with the loss function defined as:
\[
\mathcal{L} = \|x_0 - f_\theta(x_t, t)\|.
\]

During inference, the reverse process begins by sampling noise from a Gaussian distribution, $p(x_T) = \mathcal{N}(x_T; \mathbf{0}, \mathbf{I})$, and iteratively denoising it using $p_{\theta}(x_{t-1}|x_t)$ until $x_0$ is obtained. DDIM~\citep{ddim} refines this process by ensuring its marginal distribution matches that of DDPM. Consequently, during generation, only a subset of diffusion steps $\{\tau_1, \dots, \tau_S\}$ is sampled, significantly reducing inference latency.

\subsection{Basics for VAE} \label{sec:app_vae}

Variational Autoencoders (VAEs)~\cite{kingma2022autoencodingvariationalbayes} are a class of generative models that combine probabilistic reasoning with neural networks to learn the underlying distribution of high-dimensional data. A VAE consists of two components: an encoder and a decoder. The encoder maps input data $x$ to a latent variable $z$ characterized by a probabilistic distribution $q(z|x)$, typically parameterized as a Gaussian. The decoder reconstructs the input by sampling from the latent space and generating data through $p(x|z)$.

To ensure that the latent space conforms to a structured prior distribution, typically a standard Gaussian $p(z) = \mathcal{N}(0, I)$, VAEs optimize the Evidence Lower Bound (ELBO):
\[
    \mathcal{L} = \mathbb{E}_{q(z|x)}[\log p(x|z)] - D_{\mathrm{KL}}(q(z|x) \| p(z)),
\]
where the first term represents the reconstruction loss, ensuring that the generated data resembles the input, and the second term is the Kullback-Leibler divergence, which regularizes the latent space.

A key point of VAEs is the reparameterization trick, which facilitates gradient-based optimization by expressing the latent variable $z$ as:
\[
    z = \mu + \sigma \cdot \epsilon, \quad \epsilon \sim \mathcal{N}(0, I),
\]
where $\mu$ and $\sigma$ are outputs of the encoder network. 

VAEs have found applications in areas such as image synthesis, data compression, and representation learning due to their ability to generate diverse, high-quality samples while maintaining interpretability of the latent space. In our work, we employ a VAE as the backbone model and introduce two submodules to decouple the video latent representation effectively.

%% file: tables/scale.tex
\begin{table*}[t]
\begin{center}
\tabcolsep=0.2cm
% \footnotesize
\caption{Settings and performance of \modelname at different scales.
}
\vspace{-0mm}
	\label{tab:scale}
	\begin{tabular}{l||cccc|cc}
		\toprule[1.5pt]
	    Models & Depth & Num. Heads & Dim. Hidden & Num. Params.& PSNR & SSIM \\
		\midrule
           \modelname$\rm{_{small}}$ & 12 & 8 & 512 & 126M &  24.83 & 0.683 \\
           \modelname$\rm{_{base}}$ & 16 & 12 & 768 & 335M & 26.13 & 0.732  \\
           \modelname$\rm{_{large}}$ & 16 & 12 & 1536 & 1.3B & 27.16 & 0.751\\
       
		\bottomrule[1.5pt]
	\end{tabular}
\end{center}
\end{table*}

%% file: tables/baselines_app.tex
\begin{table}
    \centering
        \caption{Comparison with other concurrent works.}
    \setlength\tabcolsep{5pt}
\begin{tabular}{c||cc}
		\toprule[1pt]
	    Models~(Comp. Rate) & PSNR$\uparrow$ & LPIPS$\downarrow$ \\
        \midrule

                    CV-VAE (0.53\%) & $28.06$ & $0.24$   \\
            OD-VAE (0.53\%)     & $29.18$ & $0.19$  \\
            Open-Sora (0.53\%)& $29.89$ & $0.15$ \\
            CogVideoX (0.53\%) & $\bf{31.92}$ & $\bf{0.09}$   \\
		\midrule
            VidTwin (0.20\%)     & $28.14$ & $0.24$ \\
            VidTwin (0.48\%)     & $\bf{30.04}$ & $\bf{0.15}$ \\           
		\bottomrule[1pt]
	\end{tabular}

    \label{tab:app_baselines}
\end{table}

%% file: tables/ratio.tex
\begin{table}
    \centering
        \caption{The reconstruction quality of different compression rates.}
    \setlength\tabcolsep{8pt}
\begin{tabular}{c||cc}
		\toprule[1pt]
	    Compression Rate & PSNR$\uparrow$ & LPIPS$\downarrow$ \\
		\midrule
            0.11\%     & $24.41$ & $0.35$ \\
            0.16\%     & $27.03$ & $0.28$ \\
            0.20\%     & $28.14$ & $0.24$ \\
            0.48\%    & $\bf{30.04}$ & $\bf{0.15}$ \\            
		\bottomrule[1pt]
	\end{tabular}

    \label{tab:compr_ration}
\end{table}

%% file: tables/other.tex
\begin{table}
    \centering
        \caption{Subjective evaluation of VidTwin with increased frames and higher FPS.}
\setlength\tabcolsep{5pt}
\begin{tabular}{l||ccc}
		\toprule[1pt]
	    Model & Sem. $\uparrow$ & Tempo. $\uparrow$ & Deta. $\uparrow$ \\
		\midrule
            VidTwin     & $4.71$ & $4.62$ & $4.73$ \\
            w/ 32 frames     & $4.70$ & $4.53$ & $4.69$\\
            w/ 40 fps    & $4.73$ & $4.64$ &$4.71$ \\           
		\bottomrule[1pt]
	\end{tabular}

    \label{tab:other}
\end{table}

%% file: tables/settings.tex
\begin{table*}[h!]
\centering
\tabcolsep=0.3cm
\caption{Recommended settings for latent sizes.}
\label{tab:latent_sizes}
\begin{tabular}{ccc}
		\toprule[1.5pt]
Setting & Structure Latent & Dynamics Latent \\ 
\midrule
1 & $h_{\boldsymbol{S}} = w_{\boldsymbol{S}} = 7, n_q = 16, d_{\boldsymbol{S}} = 4$ & $h_{\boldsymbol{D}} = w_{\boldsymbol{D}} = 7, d_{\boldsymbol{D}} = 8$ \\ 
\midrule
2 & $h_{\boldsymbol{S}} = w_{\boldsymbol{S}} = 7, n_q = 16, d_{\boldsymbol{S}} = 4$ & $h_{\boldsymbol{D}} = w_{\boldsymbol{D}} = 4, d_{\boldsymbol{D}} = 16$ \\ 
\midrule
3 & $h_{\boldsymbol{S}} = w_{\boldsymbol{S}} = 7, n_q = 12, d_{\boldsymbol{S}} = 4$ & $h_{\boldsymbol{D}} = w_{\boldsymbol{D}} = 7, d_{\boldsymbol{D}} = 8$ \\ 
		\bottomrule[1.5pt]
\end{tabular}
\end{table*}

%% file: tables/hyper.tex
\begin{table}[h!]
\centering
\caption{Training Configuration}
\label{tab:training_config}
\begin{tabular}{l|l}
\toprule[1.5pt]
\textbf{Parameter}              & \textbf{Value} \\ \midrule
Input Video Resolution          & 224            \\ \midrule
Input Video Frames              & 16             \\ \midrule
Input Video FPS                 & 8              \\ \midrule
Optimizer                       & Adam; $\beta_1 = 0.9, \beta_2 = 0.99$ \\ \midrule
Learning Rate                   & $1.6 \times 10^{-4}$ \\ \midrule
Warmup Steps                    & 5000           \\ \midrule
Learning Rate Scheduler         & Cosine Annealing \\ \midrule
$\mathcal{L}_{p}$ & 0.05  \\ \midrule
Weight Decay & 0.0001 \\ \midrule
$\mathcal{L}_{GAN}$ & 0.05  \\ \midrule$\mathcal{L}_{KL}$ & 0.001  \\ \midrule
Training Batch Size             & 6              \\ \midrule
Training Device                 & 4 $\times$ 80G A100 GPUs \\ 
\bottomrule[1.5pt]
\end{tabular}
\end{table}